\pdfoutput=1

\documentclass[11pt]{article}

\usepackage[final]{coling}

\usepackage{times}
\usepackage{latexsym}

\usepackage[T1]{fontenc}

\usepackage[utf8]{inputenc}

\usepackage{microtype}

\usepackage{inconsolata}

\usepackage{graphicx}
\usepackage{amsmath}
\usepackage{amssymb}
\usepackage{color}
\usepackage{multirow}
\usepackage{float}
\usepackage{makecell}
\usepackage{booktabs} 
\usepackage{stfloats}
\usepackage{shadowtext}
\usepackage{soul} 
\usepackage{color, xcolor}
\usepackage{diagbox}

\title{Positive Text Reframing under Multi-strategy Optimization}

\author{Shutong Jia\textsuperscript{1,2}, Biwei Cao\textsuperscript{1}, Qingqing Gao\textsuperscript{1}, Jiuxin Cao\textsuperscript{1}\thanks{\ \ Corresponding author.}, Bo Liu\textsuperscript{1} \\
        \textsuperscript{1}Southeast University,
        \textsuperscript{2}State Grid Tianjin Power Dongli Power Supply Branch
        \\ 
        \texttt{\{shutong\_jia,caobiwei,qingqing\_gao,jx.cao,bliu\}@seu.edu.cn} \\
        }

\begin{document}
\maketitle
\begin{abstract}
Differing from sentiment transfer, positive reframing seeks to substitute negative perspectives with positive expressions while preserving the original meaning. With the emergence of pre-trained language models (PLMs), it is possible to achieve acceptable results by fine-tuning PLMs. Nevertheless, generating fluent, diverse and task-constrained reframing text remains a significant challenge. To tackle this issue, a \textbf{m}ulti-\textbf{s}trategy \textbf{o}ptimization \textbf{f}ramework (MSOF) is proposed in this paper.  Starting from the objective of positive reframing, we first design positive sentiment reward and content preservation reward to encourage the model to transform the negative expressions of the original text while ensuring the integrity and consistency of the semantics. Then, different decoding optimization approaches are introduced to improve the quality of text generation. Finally, based on the modeling formula of positive reframing, we propose a multi-dimensional re-ranking method that further selects candidate sentences from three dimensions: strategy consistency, text similarity and fluency. Extensive experiments on two Seq2Seq PLMs, BART and T5, demonstrate our framework achieves significant improvements on unconstrained and controlled positive reframing tasks.
\end{abstract}

\section{Introduction}

The concept of style transfer initially emerges within the domain of computer vision (CV) with the objective of accomplishing image style transfer \citep{Gatys_2016_CVPR}. Inspired by this, \citet{Hu-2018} proposed text style transfer (TST), whose main purpose is to automatically control the text style and preserve the style-independent content. 
In recent years, there has been an increasing focus on TST, which has gradually evolved into a significant subfield within the domain of natural language generation. Many corresponding task variants also have been proposed, such as text form transfer \citep{briakou-etal-2021-ola}, topic transfer \citep{huang-etal-2020-cycle}, text simplification \citep{cao-etal-2020-expertise}, and sentiment transfer \citep{mueller2017sequence}, etc.
\begin{figure}[t]
\centering
\centerline{\includegraphics[scale=0.42]{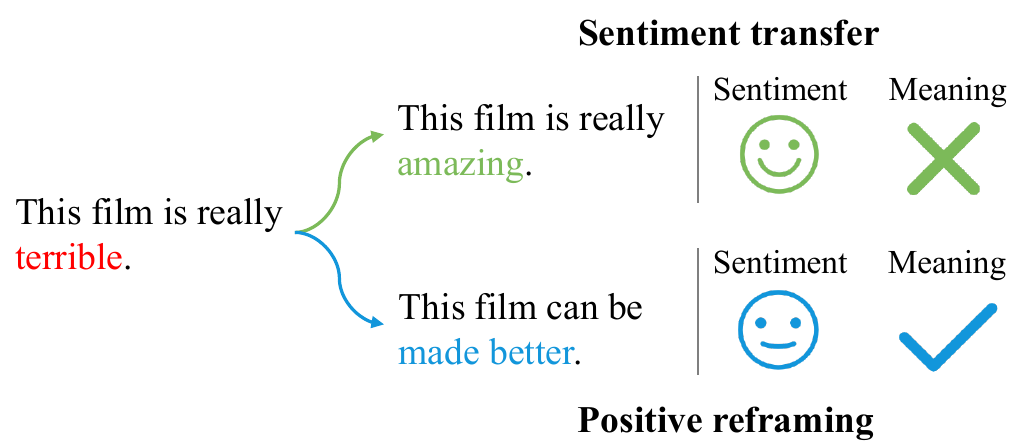}}
\caption{The difference between sentiment transfer and positive reframing.}
\label{diff}
\end{figure}

Among them, sentiment transfer primarily focuses on reversing the sentiment polarity of the original text. However, it relies on the straightforward replacement of opinion words, such as substituting negative opinion words with their positive counterparts of the opposite meaning.  On the one hand, it retains the content irrelevant to style to some extent, such as the invariance of described object entities. On the other hand, it also inherently alters the meaning of the original text \citep{liao-etal-2018-quase,li-etal-2018-delete}. To this end, \citet{ziems-etal-2022-inducing} proposed positive reframing. In contrast to sentiment transfer, positive reframing adopts principles from psychology to reframe negative text by introducing a complementary positive viewpoint while simultaneously maintaining the underlying meaning conveyed in the original text. A toy example of their difference can be seen in Figure~\ref{diff}.

More specifically, positive reframing encompasses various tasks, including unconstrained positive reframing, controlled positive reframing,  and derivative tasks such as reframe strategy classification. The unconstrained positive reframing task focuses on generating reframed text without explicit guidance of the corresponding reframe strategy. In contrast, the controlled positive reframing task involves reframing text based on the given strategy. And the reframe strategy classification task entails determining the specific strategy employed in reframing text. \citet{ziems-etal-2022-inducing} gives six positive reframing strategies, namely growth mindset, impermanence, neutralization, optimism, self-affirmation and thankfulness.

However, most of the existing methods only fine-tune PLMs on the corresponding dataset, ignoring the consistency requirement between the model training objective and the target of positive reframing, and also failing to fully utilize the known condition of the reframing strategy under the controlled setting, making it difficult to ensure that the generated text meets the task requirements.  Therefore, this paper proposes a multi-strategy optimization framework (MSOF) for positive reframing and our contributions are as follows:

$\bullet$ Firstly, from the target of positive reframing, we design and implement the positive sentiment reward and content preservation reward to optimize the sequence-level training objective, and then apply various decoding improvement approaches to alleviate text degeneration and elevate the quality and diversity of the generated text.

$\bullet$ Secondly, we propose a multi-dimensional re-ranking approach based on the modeling formula of positive reframing, which comprehensively evaluates the quality of the candidate text based on strategy consistency, text similarity and fluency.

$\bullet$ Extensive experimental results demonstrate that our proposed multi-strategy optimization framework achieves significant improvement on both unconstrained and controlled positive reframing task. And we would release our code to encourage future research\footnote{\url{https://github.com/20174376/code-for-paper}}.

\section{Related Work}
Early research on \textbf{text style transfer} mostly relied on artificial design features such as syntax \citep{carroll-etal-1999-simplifying} and phrase \citep{quirk-etal-2004-monolingual} modeling, etc. Similar to other tasks in NLP, the advent of deep learning has resulted in the growing application of neural network models to TST. For example, \citet{jhamtani-etal-2017-shakespearizing} investigated the utilization of the Seq2Seq model for transforming modern English into Shakespearean-style English. \citet{wang-etal-2019-harnessing} applied GPT-2 to accomplish the formal-informal transfer. \citet{Sancheti-2020} extended the work of \citet{jhamtani-etal-2017-shakespearizing} by incorporating a reinforcement learning framework. \citet{lai-etal-2021-thank} further applied this framework to PLMs. Above studies are mainly based on parallel corpora. Although satisfactory results can be achieved, the cost of constructing parallel corpora is expensive. Therefore, semi-supervised learning and unsupervised learning are widely used in TST. The main methods include data augmentation or text retrieval \citep{zhang-etal-2020-parallel-corpus, jin-etal-2019-imat}, adversarial learning \citep{Hu-2018, fu2018style}, back-translation \citep{prabhumoye-etal-2018-style, wei-etal-2023-text}, and reinforcement learning \citep{Luo-2019, gong-etal-2019-reinforcement}.

Specific to \textbf{sentiment transfer}, the early goal is to extract sentiment words that describe the corresponding entities, and then replace them with expressions of the opposite sentiment attribute. The representative one is the “Delete, Retrieve, Generate” strategy \citep{li-etal-2018-delete}. Furthermore, \citet{sudhakar-etal-2019-transforming} applied the transformer architecture to the above strategy. To better distinguish content and style, \citet{kim-sohn-2020-positive} divided the model into sentence reconstruction module and style module to complete their respective task. \citet{han-etal-2023-text} introduced the adaptive clustering and contrastive learning modules to better explore sentence transmission patterns to main and utilize the latent transfer patterns. 

Although sentiment transfer preserves attribute-independent content, the intrinsic meaning of the original text expression is also changed. To this end, \citet{ziems-etal-2022-inducing} introduced \textbf{positive reframing},  aiming to preserve the original meaning by substituting negative viewpoints with complementary positive expressions, and constructed the corresponding parallel dataset. For unconstrained positive reframing,
\citet{Sheng-Decoupling} decoupled the sentiment and style of the text to complete the positive reframing. Then, \citet{sheng-etal-2023-learning} further decomposed positive reframing into paraphrase generation and sentiment transfer and constructed corresponding pseudo datasets to fuse generation capabilities through multi-task learning, but also led to the inability to apply their method under the controlled setting. 

\section{Methodology}
\subsection{Problem Definition}
Let ($x$, $y$, $\psi_x$) be a triple in the positive reframing task, where $x$ = \{$x_1, x_2,\dots, x_n$\} is the original text with negative sentiment, and $y$ = \{$y_1, y_2,\dots, y_m$\} is the target sentence with complementary positive expressions corresponding to $x$, $m$ and $n$ represent the sentence length. $\psi_x$ $\subseteq$ \{Growth Mindset, Impermanence, Neutralizing, Optimism, Self-affirmation, Thankfulness\} is the positive reframing strategy used to reframe the negative text $x$, which can use multiple strategies simultaneously. This paper researches the following three tasks.

The target of \textbf{unconstrained positive reframing} is to generate the target sentence $y$ from the original text $x$ \textbf{without} any reframe strategy guidance. This task can be modeled as follows:
\begin{equation}
\begin{aligned}
p(y|x) = \prod \limits_{t=1}^m p(y_t|x, y_{<t})\label{unequ}
\end{aligned}
\end{equation}
where $y_{<t}$ represents what has been generated before time $t$.

Regarding \textbf{reframe strategy classification}, its requirement is to predict the reframing strategy $\psi_x$ used to reframe the original sentence $x$.

For \textbf{controlled positive reframing}, the primary objective is to generate the target sentence $y$ from the original text $x$ \textbf{under} given strategy $\psi_x$, This problem can be modeled as the following formula.
\begin{equation}
\begin{aligned}
p(y|x,\psi_x) = \prod \limits_{t=1}^m p(y_t|x,\psi_x,y_{<t}) \label{equ2}
\end{aligned}
\end{equation}
\begin{figure*}[ht]
\centering
\centerline{\includegraphics[width=0.9\textwidth]{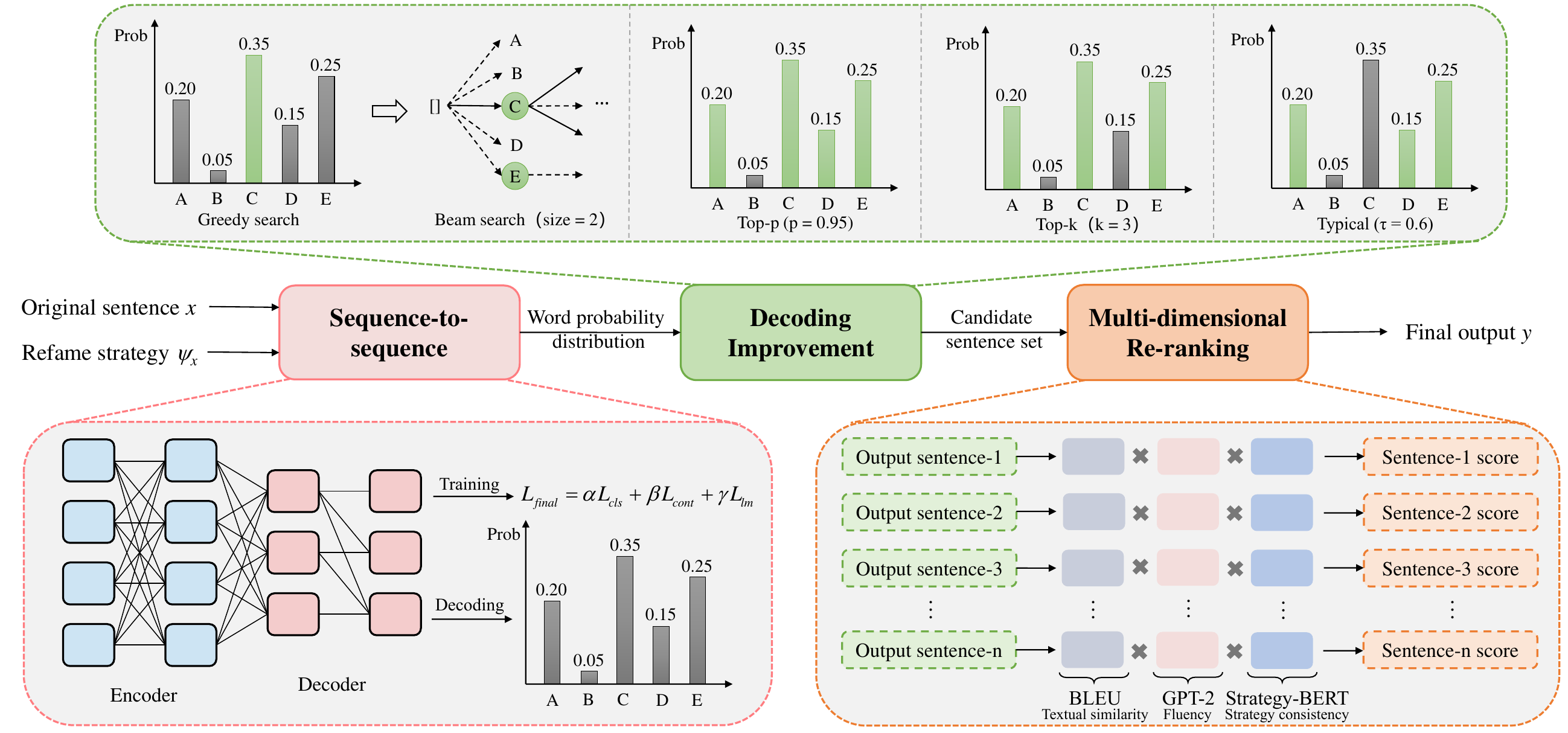}}
\caption{The overall architecture of MSOF. We respectively use BART and T5 as the basic model for positive reframing. The positive sentiment reward and content preservation reward are applied to optimize the model training process. Then, we adopt various decoding improvement approaches (e.g. beam search, random sampling) during the decoding stage to improve the quality of text generation. Finally, multi-dimensional re-ranking is used to comprehensively evaluate candidate sentences and select the candidate with the highest score as the final output.}
\label{fig2}
\end{figure*}
\subsection{Framework}
As shown in Figure~\ref{fig2}, our proposed framework mainly consists of four modules, namely sequence-to-sequence, reinforcement training, decoding improvement and multi-dimensional re-ranking.

\subsubsection{Sequence-to-sequence}
Consistent with \citet{ziems-etal-2022-inducing}, we also use T5 \citep{Raffel-2019} and BART \citep{lewis-etal-2020-bart} as the basic text generation model, which are both mainly composed of two components, namely encoder and decoder. 

\textbf{Encoder} This part is to encode original sentence $x$ and reframe strategy $\psi_x$ into hidden vector $H$. We use T5 and BART as the basic generation model, and the encoder part is as follows:
\begin{equation}
\begin{aligned}
H={\text {Encoder}}([x_1,x_2,\dots,x_n],\psi_x)
\end{aligned}
\end{equation}
where $H \in \mathbb{R}^{l \times d}$, $l$ is the length of sequence, and $d$ is the hidden dimension.

\textbf{Decoder} The output $y_t$ of the decoder part takes the hidden vector output of the encoder and the output $y_{<t}$ of the decoder before time $t$ as input, the equation is as follows.
\begin{equation}
\begin{aligned}
y_t={\text{Decoder}}(H;y_{<t}) \label{equ4}
\end{aligned}
\end{equation}
\subsubsection{Reinforcement Training}
\begin{figure}[h]
\centering
\centerline{\includegraphics[scale=0.3]{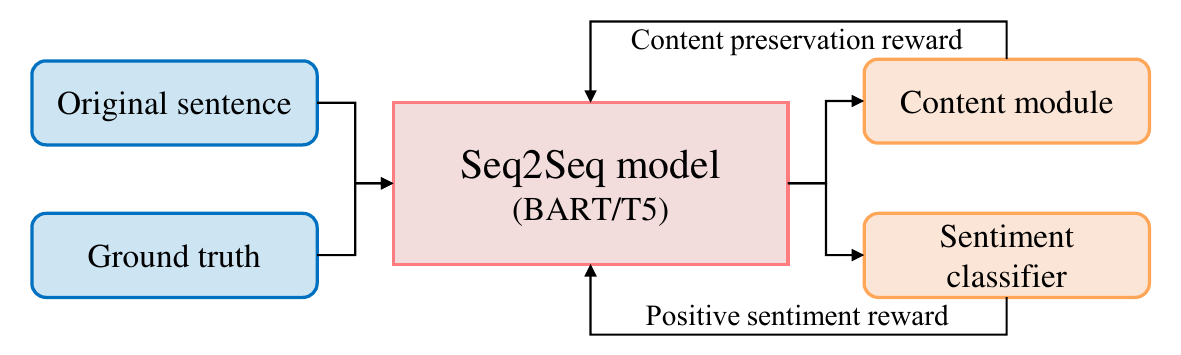}}
\caption{The reinforcement training procedure of the Seq2Seq-based model.}
\label{reward}
\end{figure}

As shown in Figure \ref{reward}, based on the objective of positive reframing, the generated text should transform the negative sentiment of the original text and keep the semantics unchanged. Therefore, we design and implement positive sentiment reward and content preservation reward to optimize the overall training process.

\textbf{Positive sentiment reward} We first design the positive sentiment reward loss based on binary cross entropy (BCE). Specifically, we fine-tune the binary sentiment classifier RoBERTa \citep{Liu-Roberta} and utilize it to determine the sentiment change degree of the generated sentence relative to the original text. The positive sentiment reward loss function is formulated as follows:
\begin{gather}
p(s_t |y',x) = {\rm Sigmoid}({\rm RoBERTa}(y',x)) \\
L_{cls} =  -{\rm log}(p(s_t |y',x))
\end{gather}
where $s_t$ represents the target style, and $y'$ is the generated sentence.

\textbf{Content preservation reward} Inspired by \citet{lai-etal-2021-thank}, we use BLEU score as the reward for content preservation and leverage SCST (Self-Critic Sequence Training)  approach \cite{Rennie_2017_CVPR} as the optimization method. The corresponding loss function is as follows:
\begin{align}
L_{cont} =\sum_{i} log(p(y_{i}^s|y_{1:i-1}^s,x))  (bleu(y',y)&\nonumber \\
- bleu(y^s, y))
\end{align}
where $y^s$ is sampled from the distribution of model outputs at each time step, and $y'$ is the greedy generation from the model.

The overall loss is a weighted sum of the positive sentiment reward loss $L_{cls}$, content preservation reward loss $L_{cont}$, and language modeling loss $L_{lm}$.
\begin{gather}
L_{lm} = \sum_{i} log(p(y_i|y_{1:i-1},x)) \\
L_{final} = \alpha L_{cls} + \beta L_{cont}  + \gamma L_{lm}
\end{gather}
\subsubsection{Decoding Improvement}
Although T5 and BART have demonstrated their superiority in the field of NLG, the sentences generated by default greedy search often result in text degeneration (i.e., empty or repeated sequences) during the decoding stage \citep{fan-etal-2018-hierarchical, Holtzman-2019}. Therefore, in this paper, various decoding improvement ways such as Beam search \citep{wiseman-rush-2016-sequence}, Top-k sampling \citep{fan-etal-2018-hierarchical}, Top-p sampling \citep{Holtzman-2019} and Typical sampling \citep{meister2023locally} are applied to the decoding stage of the Seq2Seq model to improve the quality of text generation. And Eq. \ref{equ4} is changed as follows.
\begin{align}
 y_t= {\text{Post-Processing}}({\text{Decoder}}(H;y_{<t}))
\end{align}

\subsubsection{Multi-dimensional Re-ranking}
According to Bayes Rule, we can decompose Eq.~\ref{equ2} into the product of three probabilities:
\begin{gather}
p(y|x,\psi_x) = p(\psi_x|y,x) \times p(x|y) \times p(y)
\end{gather}
The first term $p(\psi_x|y,x)$ can be seen as the consistency of original-to-generative sentence transformation with given reframe strategy\footnote{For unconstrained setting, Eq.\ref{unequ} can be decoupled as follows: $p(y|x) = p(x|y) \times p(y)$. Therefore, there is no strategy consistency evaluation.
}.
The second term $p(x|y)$ represents the textual similarity. And the last term $p(y)$ can be regarded as the overall fluency of the output.

\textbf{Strategy consistency} For this term, we propose Strategy-BERT to evaluate the consistency between text reframing and the given strategy, which draws on the idea of "breaking the whole into pieces" and prompt learning to transform the multi-label problem into multiple binary classification tasks, i.e. training the corresponding model for each reframing strategy. For one thing, this approach enables each model to concentrate on its specific aspect and thus not affect each other. For another thing, it facilitates context semantic enhancement by constructing an auxiliary sentence that incorporates supplementary task prompt to effectively mine the implicit task-specific knowledge contained in PLMs and alleviate the task awareness challenge.

\begin{figure}[h]
\centering
\centerline{\includegraphics[scale=0.21]{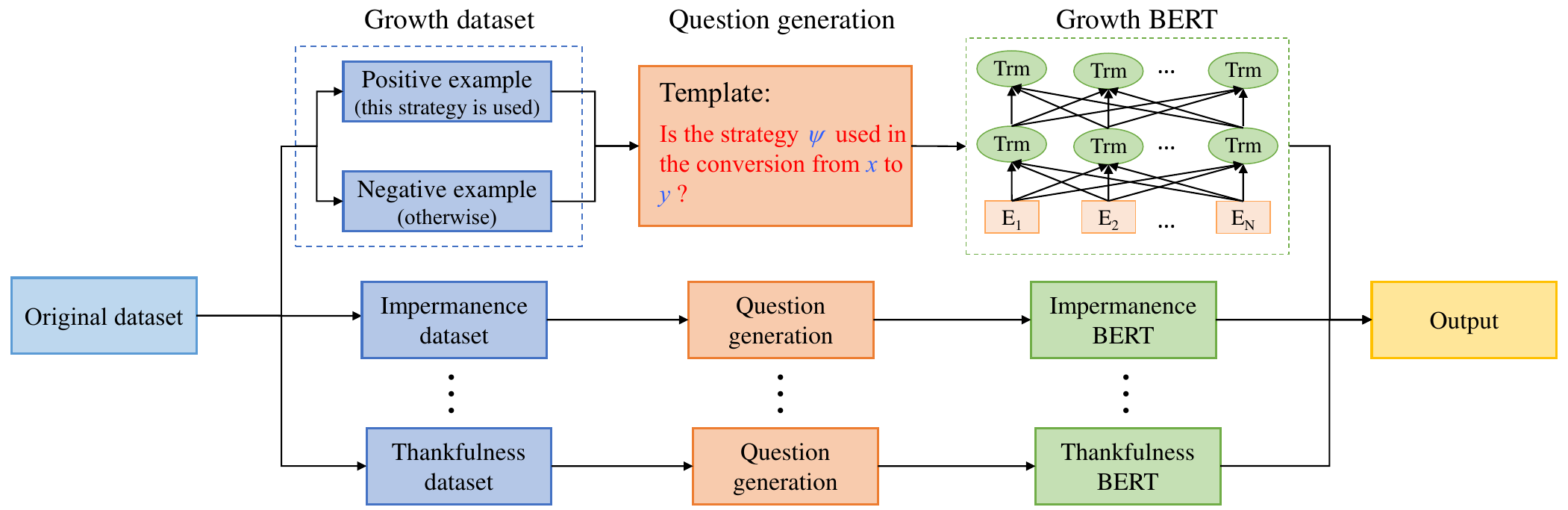}}
\caption{The overall procedure of reframe strategy classification.}
\label{fig3}
\end{figure}

As shown in Figure~\ref{fig3}, the original dataset is firstly divided according to the different strategies used in reframing, that is, if the strategy $\psi$ is used in the original-reframed text transfer, this sentence pair will be regarded as a positive sample of corresponding strategy dataset, otherwise, it will be a negative sample. The dataset division results are shown in Table \ref{tab5}.

For different reframe strategies, this paper uses the following way to construct auxiliary question:

"\textcolor{red}{Is the strategy} + \textcolor{blue}{strategy type} + \textcolor{red}{used in the conversion from} + \textcolor{blue}{original} + \textcolor{red}{to} + \textcolor{blue}{reframe} + \textcolor{red}{?}"  
where the artificially added tokens are marked in red, and the reframe strategy, original sentence and reframed sentence are marked in blue. In this way, context semantic enhancement can be achieved by constructing auxiliary question.

Then, we fine-tune BERT on above dataset and propose Strategy-BERT specific to each reframe strategy, which is used to evaluate the strategy consistency score of candidate sentences. For each candidate sentence, we invoke the corresponding evaluation model to calculate its consistency score on the strategies used in positive reframing.

\textbf{Textual similarity} We still use BLEU to calculate this term because it can ensure that the generated text preserves style-independent content \cite{Sancheti-2020}.

\textbf{Fluency} Recent works suggest that the probability of output generated from PLM is an appropriate automatic and referenceless measure of fluency \cite{suzgun-etal-2022-prompt, ramirez-etal-2023-controllable}. Therefore, we use GPT-2$_{{\rm large}}$ \cite{radford2019language} to calculate the overall fluency of each candidate.

Finally, we take the product of scores from the above three items as the final score of the candidate sentence and choose the one with the highest score as the final output.


\section{Experiment}
\subsection{Dataset}

\textbf{Reframe strategy classification} To verify the effectiveness of Strategy-BERT, we conduct experiments on reframe strategy classification task. Since this paper converts the multi-label classification problem into multiple binary classification tasks, the dataset is also divided accordingly, and the division results are presented in Table \ref{tab5}.

\textbf{Positive reframing} For unconstrained positive reframing and controlled positive reframing, we adopt the dataset provided by 
\citet{ziems-etal-2022-inducing}
. and the specific statistics are given in Table \ref{tab4}.

\begin{table}[h]
\centering
\resizebox{\linewidth}{!}{
\begin{tabular}{ccccccc}
\toprule
\multirow{2}{*}{\textbf{Label}} & \multicolumn{2}{c}{\textbf{Train}} & \multicolumn{2}{c}{\textbf{Dev}} & \multicolumn{2}{c}{\textbf{Test}}\\
\cmidrule(lr){2-3} \cmidrule(lr){4-5} \cmidrule(lr){6-7}& POS & NEG & POS & NEG & POS & NEG \\
\midrule
Growth           & 1683 & 4996 & 216 & 619 & 221 & 614\\
Impermanence     & 1296 & 5383 & 172 & 663 & 157 & 678\\
Neutralizing     & 2410 & 4269 & 303 & 532 & 302 & 533\\
Optimism         & 3295 & 3383 & 373 & 462 & 400 & 435 \\
Self-affirmation & 673  & 6006 & 92  & 743 & 76  & 759\\
Thankfulness     & 882  & 5797 & 94  & 741 & 109 & 726\\
\bottomrule
\end{tabular}}
\caption{The statistics of the reframe strategy classification dataset. 
}
\label{tab5}
\end{table}

\begin{table}[htbp]
\small
\centering
\begin{tabular}{cccc}
\toprule
\textbf{Label}  & \textbf{Train} & \textbf{Dev} & \textbf{Test}\\
\midrule
Growth & 1683 & 216 & 221\\
Impermanence & 1296 & 172 & 157 \\
Neutralizing & 2410 & 303 & 302 \\
Optimism & 3295 & 373 & 400 \\
Self-affirmation & 673 & 92 & 76\\
Thankfulness & 882 & 94 & 109 \\
\bottomrule
\end{tabular}
\caption{The statistics of the positive reframing dataset (unconstrained \& controlled).}
\label{tab4}
\end{table}

\subsection{Evaluating Metrics}
Regarding classification task, following \citet{ziems-etal-2022-inducing}, we use F1 score as the evaluation metric.

For generation task, the following nine automatic metrics are used: (1) Content preservation-related metric, namely ROUGE-1 (R-1), ROUGE-2 (R-2), ROUGE-L (R-L) \citep{lin-2004-rouge}, BLEU \citep{papineni-etal-2002-bleu} and BERTScore (BScore) \citep{Zhang-2019-BScore}. (2) $\Delta$TextBlob ($\Delta$TB) \citep{Loria-blob} is used to report the average change in sentiment. (3) RTQE (\textbf{R}eframing \textbf{T}ext \textbf{Q}uality \textbf{E}valuation) is proposed to evaluate the degree of positive text reframing (i.e. style strength), we fine-tune RoBERTa$_{{\rm large}}$  \citep{Liu-Roberta} to evaluate reframing degree and we regard the probability from the model prediction as the degree of positive reframing between the original and generated sentence; on the human reference it has the F1 score of 95.98\% and accuracy of 97.41\%. For more details refer to Appendix \ref{fir:Evaluation Model} (4) Perplexity (PPL) is an indicator of text fluency, and we use GPT-2$_{{\rm large}}$ as the evaluation model. 

Finally, following \citet{ziems-etal-2022-inducing}, we randomly selected 50 samples from each generated file and assigned them to 3 well-educated raters with relevant professional backgrounds to score Meaning Preservation (Meaning), Positivity and Fluency of reframed sentences on a scale of 1 to 5.
Since the main research of this paper falls on controlled positive reframing task, we only conducted human evaluation on this task. 

\subsection{Implementation Details}
\label{fir:Implement Details}

\textbf{Reframe strategy classification} BERT$_{{\rm base}}$ \citep{devlin-etal-2019-bert} and RoBERTa$_{{\rm base}}$ \citep{Liu-Roberta} are used as the backbone model in this task respectively. The maximum text embedding length is set to 110. AdamW is used as the optimizer, and the batch size is 16. In addition, all models in this paper are implemented through HuggingFace \citep{wolf-etal-2020-transformers} and PyTorch \citep{paszke2019pytorch} on TITAN Xp GPU.


\textbf{Positive reframing} Following \citet{ziems-etal-2022-inducing},  we use T5 \citep{Raffel-2019} and BART \citep {lewis-etal-2020-bart} with 6 layers in each of the encoder and decoder, and the hidden size of 768. The value of the learning rate is from 3e-5 to 3e-4, the batch size processed by each device is 6, and the text maximum input length is 80. $\alpha$, $\beta$, $\gamma$ are respectively set to 1, 0.2, 1,  which are the choices obtained from multiple experiments. And the approach of obtaining the candidate sentence set can be seen in Appendix \ref{fir:candidate sentence}.

\subsection{Main Results}
\label{fir:Main Results}

\subsubsection{Reframe Strategy Classification}
For this task, this paper selects the Multi-label-BERT and Multi-label-RoBERTa proposed by \citet{ziems-etal-2022-inducing} as baselines to compare with the Strategy-BERT and Strategy-RoBERTa proposed in this paper. For fairness, we directly adopt the results reported by \citet{ziems-etal-2022-inducing}. Since they only report F1 score of their models, we only use it as the evaluation metric in this task. The detailed performance of our proposed models on other metrics can be found in Table \ref{detailclass} in Appendix \ref{sec:Reframe strategy classification}. 

\begin{table}[htbp]
\centering
\resizebox{\linewidth}{!}{
\begin{tabular}{ccc|cc}
\toprule
\textbf{Label}  & \makecell[c]{\textbf{Multi-label-} \\ \textbf{BERT}} & \makecell[c]{\textbf{Multi-label-} \\ \textbf{RoBERTa}} &  \makecell[c]{\textbf{Strategy-} \\ \textbf{BERT}} & \makecell[c]{\textbf{Strategy-} \\ \textbf{RoBERTa}} \\
\midrule
Thankfulness & 0.71 & 0.69 & \textbf{0.73} & 0.72\\
Neutralizing & 0.59 & 0.61 & \textbf{0.61} & 0.61 \\
Optimism     & 0.71 & 0.71 & 0.71 & \textbf{0.73}  \\
Impermanence & 0.55 & 0.55 & \textbf{0.57} & 0.57  \\
Growth & 0.63 & 0.63 & 0.67 & \textbf{0.69}  \\
Self-affirmation & 0.43 & 0.44 & \textbf{0.48} & 0.46  \\
\bottomrule
\end{tabular}}
\caption{The experimental results of reframe strategy classification on F1 score. And the best results in each label are in \textbf{bold}.}
\label{tab7}
\end{table}

It can be seen from Table \ref{tab7} that our models are able to outperform baselines on all labels, significantly on the Growth (Growth Mindset) label, the two models proposed in this paper have increased by 4 points and 6 points respectively. Furthermore, in terms of the Self-affirmation label, Strategy-BERT demonstrates a noteworthy improvement of 5 points compared to the corresponding baseline. Additionally, our method consistently achieves approximately 1 point of improvement on other labels, further affirming the effectiveness and superiority of our approach. Since the performance of Strategy-BERT and Strategy-RoBERTa are similar, we only use Strategy-BERT as the evaluation model to measure the strategy consistency of each candidate. 

\begin{table}[h]
\centering
\resizebox{\linewidth}{!}{
\begin{tabular}{ccc|cc}
\toprule
\textbf{Label}  & \makecell[c]{\textbf{Strategy-BERT} \\ \textbf{w/o auxiliary}} & \makecell[c]{\textbf{Strategy-} \\ \textbf{BERT}} &  \makecell[c]{\textbf{Strategy-RoBERTa} \\ \textbf{w/o auxiliary}} & \makecell[c]{\textbf{Strategy-} \\ \textbf{RoBERTa}} \\
\midrule
Thankfulness & 0.71 & \textbf{0.73} & 0.69 & 0.72\\
Neutralizing & 0.59 & \textbf{0.61} & 0.60 & 0.61 \\
Optimism     & 0.71 & 0.71 & 0.71 & \textbf{0.73}\\
Impermanence & 0.55 & \textbf{0.57} & 0.55 & 0.57  \\
Growth       & 0.61 & 0.67 & 0.65 & \textbf{0.69}  \\
Self-affirmation & 0.44 & \textbf{0.48} & 0.44 & 0.46  \\
\bottomrule
\end{tabular}
}
\caption{The experimental results of different input ways on F1 score. The best results in each label are in \textbf{bold} and w/o auxiliary means without using auxiliary sentence.}
\label{w/o auxiliary}
\end{table}

In addition, the performance of the input approach of directly connecting the original and generated sentence is also tested to demonstrate the effectiveness of the contextual semantic enhancement strategy (i.e., the construction of auxiliary question) used in this paper. And the experimental results are given in Table \ref{w/o auxiliary}. As can be seen, the F1 score on each label is greatly reduced without context enhancement strategy, but our models still achieve comparable performance with the multi-label classification models which once again shows the effectiveness of our method.

\sethlcolor{lightgray}
\begin{table*}[htbp]
 \small
\centering
\begin{tabular}{llcccccccc}
\toprule 
\multicolumn{2}{c}{\textbf{Model}} &  \textbf{R-1} & \textbf{R-2} & \textbf{R-L} & \textbf{BLEU} & \textbf{BScore} & \textbf{$\Delta$TB} & \textbf{RTQE} & \textbf{PPL}    \\ \midrule

\multirow{9}{*}{\textbf{T5}}

&Vallina Fine-tune \citep{ziems-etal-2022-inducing} & 27.4 & 9.8 & 23.8 & 8.7 & 88.7 & 0.38 & 84.8 & 42.7\\

&FDSC \cite{Sheng-Decoupling} & 30.4 & 10.9 & 25.2 & 8.1 & 88.8 & 0.39 & 93.1 & 30.0\\ 

&PG2ST \cite{sheng-etal-2023-learning} & 31.1 & 11.2 & 25.5 & 8.9 & 88.7 & 0.35 & 85.4 & 41.0\\

&ST2PG \cite{sheng-etal-2023-learning} & 30.8 & 11.3 & 25.5 & 8.8 & 88.7 & 0.33 & 84.6 & 43.2 \\

&MSOF$_{{\rm Greedy}}$ & 32.9 & 13.0 & 26.0 & 8.8 & 89.1 & 0.37 & 86.2 & 36.8\\



&MSOF$_{{\rm Beam}}$ & 34.1 & 14.0 & 27.1 & 9.7 & 89.2 & 0.37 & 89.0 & 35.4\\ 



&MSOF$_{{\rm Top-k}}$ &  
\textbf{34.8} & \textbf{14.7} & \textbf{27.7} & \textbf{10.1} & \textbf{89.5} & \textbf{\hl{0.44}} & 93.5 & 22.3\\



&MSOF$_{{\rm Top-p}}$ & 34.4 & 14.6 & 27.6 & 10.1 & 89.4 & 0.43 & 93.5 & \textbf{\hl{22.2}}\\


&MSOF$_{{\rm Typical}}$ & 32.9 & 13.5 & 26.2 & 9.1 & 89.3 & 0.39 & \textbf{\hl{94.5}} & 22.6\\


\midrule
\multirow{9}{*}{\textbf{BART}}
&Vallina Fine-tune \citep{ziems-etal-2022-inducing} & 27.7 & 10.8 & 24.3 & 10.3 & 89.3 & 0.23 & 63.8 & 86.0\\ 

&FDSC \cite{Sheng-Decoupling} & 32.7 & 13.4 & 27.0 & 10.4 & 88.5 & 0.21 & 60.1 & 77.5\\ 

&PG2ST \cite{sheng-etal-2023-learning} & 32.6 & 13.5 & 26.9 & 10.3 & 88.4 & 0.19  & 60.9 & 86.2 \\ 

&ST2PG \cite{sheng-etal-2023-learning} & 32.9 & 13.6 & 27.1 & 10.9 & 88.4 & 0.20 & 61.5 & 78.9\\

&MSOF$_{{\rm Greedy}}$ & 32.3 & 13.2 & 26.9 & 10.4 & 89.4 & 0.24 & 80.1 & 47.0 \\

&MSOF$_{{\rm Beam}}$ & 34.2 & 14.2 & 28.1 & 10.9 & 89.5 & 0.24 & 87.3 & 33.6\\




&MSOF$_{{\rm Top-k}}$ &  \textbf{\hl{34.8}} & \textbf{\hl{14.9}} & \textbf{\hl{29.3}} & \textbf{\hl{12.0}} & \textbf{\hl{89.9}} & \textbf{0.31} & 87.3 & \textbf{25.8}\\

&MSOF$_{{\rm Top-p}}$ & 34.8 & 14.9 & 29.2 & 12.0 & 89.8 & 0.30 & 87.2 & 27.3\\

&MSOF$_{{\rm Typical}}$ & 32.5 & 12.8 & 26.9 & 10.4 & 89.5 & 0.30 & \textbf{88.5} & 29.6\\ 

\bottomrule
\end{tabular}
\caption{The experimental results of \textbf{unconstrained positive reframing}.  The best in-category performance is \textbf{bolded} and the best overall performance is \textbf{\hl{highlighted}}. And except for PPL, all other metrics are better when they are higher.}
\label{unconstrained}
\end{table*}

\subsubsection{Unconstrained Positive Reframing}
As shown in Table \ref{unconstrained}, our proposed framework MSOF achieves significant improvements compared to the baselines. When combining positive sentiment reward and content preservation reward only during the training process, i.e. MSOF$_{{\rm Greedy}}$, already outperforms the baselines on almost all metrics, especially ROUGE, BScore, RTQE, and PPL. When incorporating decoding optimization and multi-dimensional re-ranking, the performance of the model will be further improved. From the perspective of the model, the T5-based models achieve the best results on metrics such as $\Delta$TB, RTQE and PPL, while the BART-based models reach SOTA on content preservation-related metrics such as ROUGE, BLEU, and BScore. This may be because BART prioritizes semantic preservation rather than sentiment change when reframing the negative text. Among different decoding methods, both beam search and random sampling-based methods are superior to greedy search. Specifically, Top-k sampling has the best overall performance, achieving the best or sub-optimal results on almost all metrics. Top-p sampling performs slightly lower than Top-k sampling. Compared to the above two decoding methods, beam search and Typical sampling are not satisfactory but still superior to the baseline method. Ultimately, regardless of whether T5 or BART is used as the basic generation model,  MSOF$_{{\rm Top-k}}$ achieves the best results among all variant models, basically achieving at least 7\% improvement on each metric compared to baselines, which strongly indicates the effectiveness of our proposed framework. 

\begin{table*}[htbp]
\centering
\small
\begin{tabular}{llccccccccc}
\toprule
 \multicolumn{2}{c}{\textbf{Model}} &  \textbf{R-1} & \textbf{R-2} & \textbf{R-L} & \textbf{BLEU} & \textbf{BScore} & \textbf{$\Delta$TB} & \textbf{RTQE} & \textbf{PPL}    \\
\midrule
\multirow{6}{*}{\textbf{T5}}
&Vallina Fine-tune \citep{ziems-etal-2022-inducing} & 27.7 & 10.0 & 23.9 & 8.8 & 88.8 & 0.36 & 86.2 & 62.1\\

&MSOF$_{{\rm Greedy}}$ & 33.6 & 13.6 & 26.7 & 8.8 & 89.2 & 0.37 & 94.6 & 34.6\\ 



&MSOF$_{{\rm Beam}}$ & 34.6 & 14.4 & 27.5 & 9.5 & 89.3 & 0.36 & 96.2 & 34.5\\  




&MSOF$_{{\rm Top-k}}$ & \textbf{34.8} & \textbf{\hl{15.0}} & \textbf{28.0} & \textbf{9.9} & \textbf{89.5} & \textbf{\hl{0.43}} & \textbf{\hl{97.7}} & 23.1\\


&MSOF$_{{\rm Top-p}}$ & 34.1 & 14.2 & 27.6 & 9.3 & 89.5 & 0.42 & 96.6 & \textbf{\hl{23.0}}\\


&MSOF$_{{\rm Typical}}$ & 33.2 & 13.4 & 26.5 & 8.6 & 89.3 & 0.42 & 97.0 & 23.8\\


\midrule
\multirow{6}{*}{\textbf{BART}}
&Vallina Fine-tune \citep{ziems-etal-2022-inducing} & 28.8 & 10.9 & 25.1 & 10.1 & 89.6 & 0.27 & 69.5 & 89.1\\

&MSOF$_{{\rm Greedy}}$ & 33.0 & 13.3 & 27.2 & 10.0 & 89.6 & 0.31 & 89.1 & 44.4\\

&MSOF$_{{\rm Beam}}$ & 34.6 & 14.2 & 28.2 & 10.5 & 89.7 & 0.34 & \textbf{94.8} & 31.8 &\\




&MSOF$_{{\rm Top-k}}$ & \textbf{\hl{34.8}} & \textbf{14.7} & \textbf{\hl{29.0}} & \textbf{\hl{11.4}} & \textbf{\hl{90.1}} & \textbf{0.36} & 94.0 & \textbf{29.4}\\

&MSOF$_{{\rm Top-p}}$ & 34.6 & 14.4 & 28.8 & 11.3 & 90.0 & 0.36 & 94.0 & 30.8\\

&MSOF$_{{\rm Typical}}$ & 33.2 & 13.2 & 27.5 & 10.1 & 89.8 & 0.36 & 94.0 & 29.8\\

\bottomrule
\end{tabular}
\caption{The experimental results of \textbf{controlled positive reframing}.}
\label{controlled exp}
\end{table*}

\subsubsection{Controlled Positive Reframing}
Since only \citet{ziems-etal-2022-inducing} have studied controlled positive reframing, we use T5 and BART \cite{ziems-etal-2022-inducing} that are fine-tuned on the corresponding dataset as baselines for comparison. The primary experimental results are given in Table \ref{controlled exp}. It can be concluded that the performance of models under constraints is generally better than unconstrained, which provides support for the reframe strategy to play a role in assisting model inference to a certain extent.  Consistent with the experimental results under the unconstrained setting, MSOF$_{{\rm Top-k}}$ still achieves the best results among all variant models. Compared with the baselines, MSOF$_{{\rm Top-k}}$ achieves an average improvement of 5 points on ROUGE, 1 point in BLEU, more than 10 points on both RTQE and PPL, and an improvement of about 20\% on $\Delta$TB. Moreover, it can be found that although Typical sampling does not perform as well as other decoding approaches on content preservation-related metrics such as ROUGE, BLEU, and BScore, it still achieves impressive results on $\Delta$TB, RTQE and PPL, suggesting that its corresponding output is consistent with task requirements to some extend, even though there is less overlap with human reference. 

\subsubsection{Ablation Experiment}

In addition, from the ablation experimental results shown in Table \ref{ablation}, we can conclude that applying content preservation reward helps the model perform well on ROUGE, BLEU and BScore, but hinders the model from transferring text style. When using positive sentiment reward, although the model performs well on $\Delta$TB and RTQE, it is not satisfactory in terms of content preservation. However, when the two are combined, the model can achieve a better balance between sentiment change and content preservation, exhibiting a more comprehensive performance. Furthermore, it can be observed that the multi-dimensional re-ranking significantly improves the model's performance on multiple metrics. This demonstrates that it can effectively select the sentence from the candidate that better meets the requirements of positive reframing. Based on the above experimental results and analysis, the validity and rationality of each component of MSOF can be effectively illustrated. For more ablation experiments, please refer to Tables \ref{unconsT5} and \ref{unconsBART} in Appendix \ref{sec:Unconstrained Positive Reframing} and Tables \ref{controT5}, \ref{controBART} and \ref{detailCon} in Appendix \ref{sec:Controlled positive reframing}.


\begin{table*}[htbp]
\centering
\small
\begin{tabular}{llccccccccc}
\toprule
 \multicolumn{2}{c}{\textbf{Model}} &  \textbf{R-1} & \textbf{R-2} & \textbf{R-L} & \textbf{BLEU} & \textbf{BScore} & \textbf{$\Delta$TB} & \textbf{RTQE} & \textbf{PPL}    \\
\midrule

\multirow{4}{*}{\textbf{T5}}

& MSOF$_{{\rm Top-k}}$ & 34.8 & \textbf{15.0} & \textbf{28.0} & \textbf{9.9} & 89.5 & \textbf{0.43} & \textbf{97.7} & \textbf{23.1}\\

& ~~ w.o Cls & 34.5 & 14.5 & 27.5 & 9.4 & 89.4 & 0.41 & 96.7 & 25.3\\

& ~~ w.o Cont & \textbf{35.0} & 14.8 & 27.7 & 9.6 & \textbf{89.6} & 0.37 & 95.7 & 24.2 \\

& ~~ w.o Re-ranking & 32.1 & 12.0 & 25.2 & 7.6 & 89.1 & 0.43 & 96.1 & 28.3\\

\midrule

\multirow{4}{*}{\textbf{BART}}

& MSOF$_{{\rm Top-k}}$ & \textbf{34.8} & \textbf{14.7} & \textbf{29.0} & \textbf{11.4} & \textbf{90.1} & 0.36 & \textbf{94.0} & \textbf{29.4}\\

& ~~ w.o Cls  & 33.6 & 13.7 & 28.2 & 10.8 & 90.0 & 0.35 & 86.9 & 31.3\\

& ~~ w.o Cont  & 33.1 & 13.7 & 27.5 & 10.9 & 89.7 & \textbf{0.38} & 86.2 & 34.6\\

& ~~ w.o Re-ranking & 31.9 & 11.9 & 26.2 & 9.4 & 89.6 & 0.35 & 92.9 & 38.8\\

\bottomrule
\end{tabular}
\caption{The ablation experimental results of MSOF under controlled setting. w.o Cls means without positive sentiment reward, w.o Cont represents without content preservation reward, w.o Re-ranking represents not using multi-dimensional re-ranking.}
\label{ablation}
\end{table*}

\subsubsection{Human Evaluation}
Finally, we adopt human evaluation to manually judge the quality of the reframed text. As can be seen from Table \ref{tabhuman}, our method is more applicable to T5, but for BART, its performance on Positivity is not satisfactory, which can also be reflected by $\Delta$TB and RTQE. Combining the relevant experimental results in Table \ref{controlled exp}, we speculate this is because the BART-based models prioritize content preservation over sentiment change. In general, consistent with the results and conclusion of automatic metrics, our method can effectively improve the model's performance, where the T5-based models perform better on Positivity and have a slightly higher score on Fluency, while BART-based models are better on Meaning.

\begin{table}[htbp]
\resizebox{\linewidth}{!}{
\centering

\begin{tabular}{lccc}
\toprule
\textbf{Model} &  \textbf{Meaning} & \textbf{Positivity} & \textbf{Fluency}\\
\midrule
T5 \citep{ziems-etal-2022-inducing} & 4.13 & 3.89 & 4.07 \\ 



MSOF$_{{\rm Top-k}}$  & \textbf{4.38} & \textbf{\hl{4.22}} & \textbf{\hl{4.58}}\\


\midrule
BART \citep{ziems-etal-2022-inducing} & 4.23 & 4.07 & 4.27 \\ 



MSOF$_{{\rm Top-k}}$  & \textbf{\hl{4.42}} & \textbf{4.10} & \textbf{4.54} \\


\bottomrule
\end{tabular}
}
\caption{The human evaluation results of controlled positive reframing. 
}
\label{tabhuman}
\end{table}

\section{Conclusion} 

We propose an original multi-strategy optimization framework (MSOF), which consists of reinforcement training, decoding improvement, and multi-dimensional re-ranking, to enhance the performance of PLMs on positive reframing. By conducting extensive experiments on T5-based and BART-based models separately, our framework achieves significant improvements over the baselines on various metrics. Future work includes further cleaning and expansion of the existing dataset to improve the quality and alleviate the imbalanced distribution of different reframe strategy labels, then exploring how the thought of controlled text generation can be applied to this task, followed by trying different approaches of context enhancement, and finally exploring how to apply large language models (LLMs) to positive reframing.

\section*{Limitations}
Firstly, the multi-strategy optimization framework proposed in this paper introduces reinforced reward in the model training stage and the multi-dimensional re-ranking to select the candidate text generated by the model. Therefore, compared with the baselines, our proposed framework needs more memory space and time during training and prediction. Then, this paper finds that the dataset provided by  \citet{ziems-etal-2022-inducing} has certain noise and label imbalance issues that may hinder the training of the model and there are currently no corresponding datasets in other languages. Finally, we also suggest that if PLMs could be further trained in a rich psychological corpus, the performance would be improved more.

\section*{Ethics Statement}
Similar to sentiment transfer, positive reframing has two sides, that is, our method can also be used to generate negative text and cause possible harmful effects on society. However, we still make our code public and hope others will be aware of the possible risks. We welcome any discussion and suggestions to minimize such risks.

\section*{Acknowledgement}
This work is supported by National Natural Science Foundation of China under Grants No.62172089, No.62472092, No.62106045. Natural Science Foundation of Jiangsu Province, China under Grants No.BK20241751. Jiangsu Provincial Key Laboratory of Computer Networking Technology, China. Jiangsu Provincial Key Laboratory of Network and Information Security, China under Grants No.BM2003201, and Key Laboratory of Computer Network and Information Integration of Ministry of Education of China under Grants No.93K-9, Nanjing Purple Mountain Laboratories, China. Start-up Research Fund of Southeast University under Grants No.RF1028623097. We thank the Big Data Computing Center of Southeast University for providing the facility support on the numerical calculations.

\bibliography{custom}
\appendix

\section{Reframing Text Quality Evaluation}
\label{fir:Evaluation Model}
\subsection{Problem Statement}
The essence of existing TST metrics such as ROUGE and BLEU is to evaluate the similarity between the generated and reference sentence, so a simple copy can lead to a high score \citep{fan-etal-2018-hierarchical,Holtzman-2019}. And for an original sentence, there may be multiple corresponding reframed sentences, especially in the unconstrained case. Furthermore, existing metrics also cannot directly measure the degree of positive reframing. Therefore, this paper proposes a new metric RTQE (\textbf{R}eframing \textbf{T}ext \textbf{Q}uality \textbf{E}valuation), which aims to evaluate the degree of positive reframing relationship between the generated and original text that can avoid the limitation of only compared with human reference given in the dataset.

\subsection{Evaluation Model}
\begin{figure}[h]
\centering
\centerline{\includegraphics[scale=0.26]{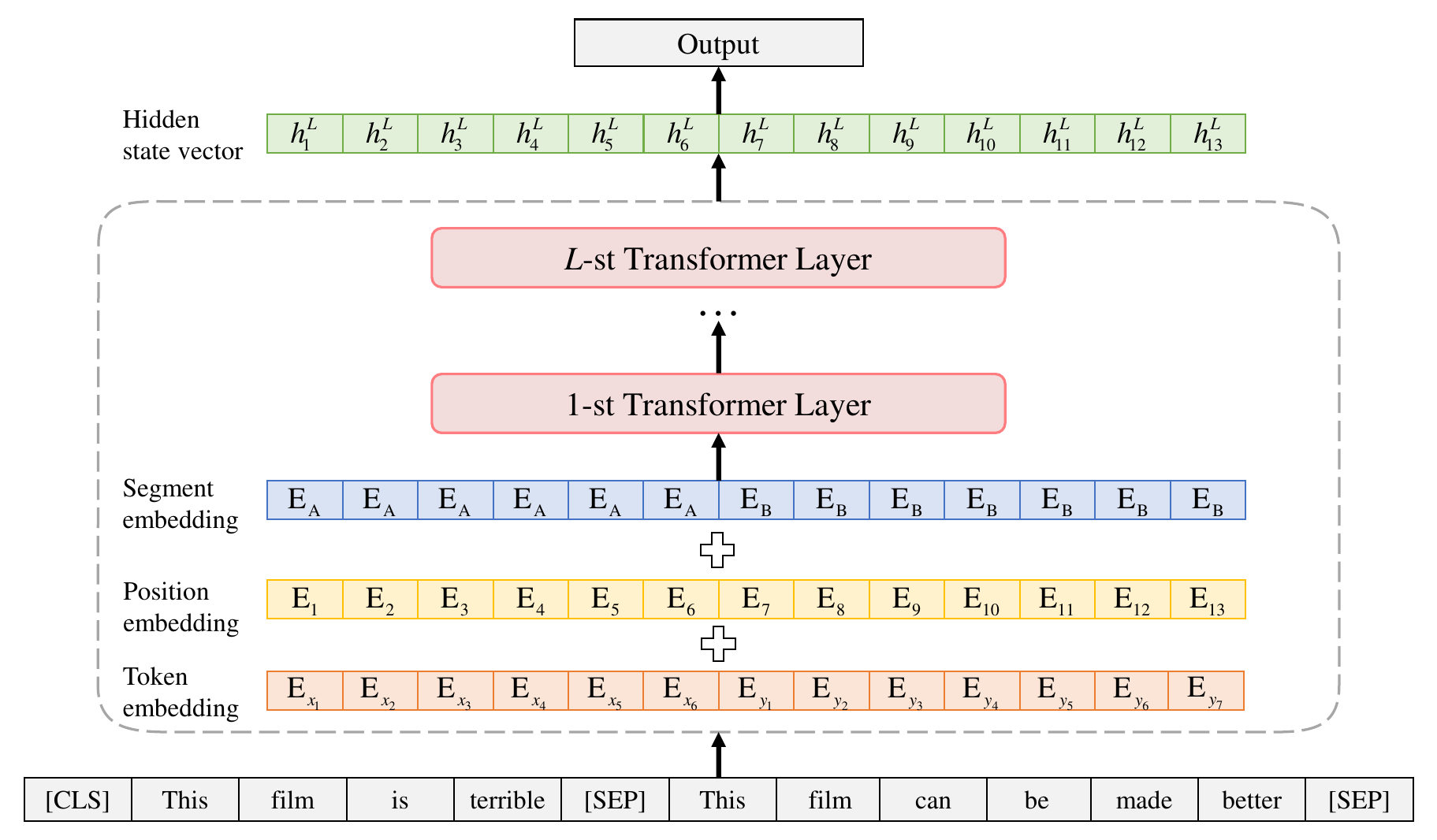}}
\caption{The model for RTQE.}
\label{Reframing Text Quality Evaluation}
\end{figure}
Taking the inspiration from \citet{lai-etal-2021-thank}, the above problem is simplified into a binary classification task, i.e., judging whether there is a positive reframing relationship between two sentences. In practical evaluation, we regard the probability from the model prediction as the degree of positive reframing between the original and generated sentence. And the RTQE evaluation model established in this paper is shown in Figure~\ref{Reframing Text Quality Evaluation}. Given the original sentence $x$ and the corresponding sentence $y$, we firstly concatenate them and input into the auto-encoding models such as BERT \citep{devlin-etal-2019-bert} and RoBERTa \citep{Liu-Roberta} (without segment embedding). The encoder part is as follows:
\begin{equation}
\begin{aligned}
H^e = {\rm \text{Encoder}}({\rm [CLS]}, x, {\rm [SEP]}, y, {\rm [SEP]}) \label{equ1}
\end{aligned}
\end{equation}
where [CLS] and [SEP] are special tokens.

The feature vector can be refined through $L$-layer transformer and the representation of $H^l$ at the $l$-th layer ($l\in$ [1, $L$]) is calculated as below:
\begin{equation}
\begin{aligned}
H^l = {\text {Transformer}}_l(H^{l-1}), H^0 = H^e 
\end{aligned}
\end{equation}
We regard the hidden vector $H^{{\rm [CLS]}}$ corresponding to [CLS] at the last layer as the contextualized representation of the whole sequence. And the prediction is obtained through the following equation:
\begin{equation}
\begin{aligned}
\text{Output} = {\rm Sigmoid}(W_{o} H^{\rm [CLS]} + b_o) \label{equ3}
\end{aligned}
\end{equation}
where $W_o \in \mathbb{R}^{{\rm dim}_H \times |y|}$is the learnable parameter of the linear layer and $b_o$ is the bias.

\subsection{Dataset}
As we simplified the RTQE task as a binary classification question, which determines whether two sentences constitute the positive reframing relationship. Therefore, this paper reconstructs the positive reframing dataset \citep{ziems-etal-2022-inducing} in the following way: for each original sentence, we consider its corresponding reframing sentence as a positive sample, and we pair the original sentence with itself or randomly select other reframing sentences to create negative samples, aiming to enhance the learning depth and generalization ability of the model. The specific statistics are presented in Table \ref{RTQEdataset}.
\begin{table}[htbp]
\small
\centering
\begin{tabular}{ccc}
\toprule
\textbf{Set} & \textbf{Positive} & \textbf{Negative}\\
\midrule
Train & 6679 & 13358 \\
Dev & 835 & 1670 \\
Test & 835 & 1670 \\
\bottomrule
\end{tabular}
\caption{The statistics of the RTQE dataset.}
\label{RTQEdataset}
\end{table}

\subsection{Implementation Details}
We use BERT \citep{devlin-etal-2019-bert} and RoBERTa \citep{Liu-Roberta} as the backbone model respectively. For the base version, the model has 12 transformer encoder layers, and the hidden size is 768. For the large version, the model has 24 transformer encoder layers, and the hidden size is 1024. In this paper, the maximum text embedding length is set to 100 tokens, AdamW with an initial learning rate 1e-5 is used as the optimizer, and batch size is 32.  

\subsection{Experiment Results}
This paper mainly tests the performance of four models: BERT$_{{\rm base}}$, BERT$_{{\rm large}}$, RoBERTa$_{{\rm base}}$ and RoBERTa$_{{\rm large}}$. And the experimental results are shown in Table \ref{tab6}.
\begin{table}[htbp]
\centering
\resizebox{\linewidth}{!}{
\begin{tabular}{cccccc}
\toprule
\textbf{Model}  & \textbf{P}(\%) & \textbf{R}(\%) & \textbf{F1}(\%) & \textbf{Acc}(\%) & \textbf{Ref}(\%)\\
\midrule
BERT$_{{\rm base}}$ & 94.49& 92.09 & 93.41 & 96.37 & 93.36\\
BERT$_{{\rm large}}$ & 95.65 & 94.85 & 95.25 & 96.85 & 93.49 \\
RoBERTa$_{{\rm base}}$ & 94.52 & 94.97 & 94.74 & 96.48 & 94.59 \\
RoBERTa$_{{\rm large}}$ & \textbf{96.16} & \textbf{96.05} & \textbf{96.11} & \textbf{97.41} & \textbf{95.98} \\
\bottomrule
\end{tabular}}
\caption{The experimental results of RTQE task. The column of Ref refers to the average degree of positive reframing relationship between the human reference and original text in the test set obtained by our models. The best results are in \textbf{bold}.}
\label{tab6}
\end{table}

It can be seen from Table \ref{tab6} that the performance of RoBERTa is generally better than BERT on all metrics, and the large version is better than the base, which shows that the more parameters and training corpus the model has, the better its performance will be. In the end, RoBERTa$_{{\rm large}}$ basically achieves the best results in all metrics and also reaches the F1 score of 95.98\% and accuracy of 97.41\% in the test of evaluating human reference, so finally this paper uses it as the evaluation model for RTQE.

Finally, we present the Pearson correlation between RTQE and manual evaluation in Table \ref{correlation}. It can be inferred that both the results of the T5-based models and BART-based models show a positive correlation with the three human evaluation metrics, particularly in terms of meaning preservation. This demonstrates that the introduction of the RTQE metric aligns with the task requirements, that is, positive reframing needs to prioritize maintaining the original meaning intact.

\begin{table}[htbp]
\centering
\resizebox{\linewidth}{!}{
\begin{tabular}{lccc}
\toprule
& \textbf{Meaning}  & \textbf{Positivity} &  \textbf{Fluency}\\
\midrule
T5-based models & 0.78 & 0.22 & 0.91 \\
BART-based models & 0.85 & 0.62 & 0.43\\
\bottomrule
\end{tabular}}
\caption{Pearson correlation between RTQE and human evaluation.}
\label{correlation}
\end{table}

\section{The Approach of Obtaining the Candidate Sentence}
\label{fir:candidate sentence}
The approach of obtaining the candidate sentence set is as follows: when beam search is used, the number of candidate sentences with the same beam size can be returned directly, and beam size of 4, 5, and 6 are experimented in this paper; for Top-k sampling, the generated sentences of $k$ = 30, 40, 50 and 60 are composed of candidate sentence set; for Top-p sampling, the generated sentences of $p$ = 0.80, 0.85, 0.90 and 0.95 are selected to be composed the candidate sentence set; for Typical sampling, the sentences generated by $\tau$ = 0.20 and 0.95 are selected according to the settings recommended by \citet{meister2023locally} to form the candidate sentence set. 

\section{The Instruction for Human Evaluation}
\label{fir:Instruction for Human Evaluation}
The specific instruction for human evaluation is as follows.

Give the \textbf{original sentence} with negative viewpoint and \textbf{reframed sentence} generated by our models. You need to score the Meaning Preservation (Meaning), Positivity and Fluency of the reframed sentence on a scale of 1 to 5.

\textbf{Meaning:} Indicate whether the reframed sentence preserves the original meaning.


\textbf{1:} Completely changed the original meaning.

\textbf{3:} Meaning related but with slight inconsistency or contradiction.

\textbf{5:} Faithful to the original meaning. 

Choose \textbf{2} or \textbf{4} when you are hesitant.

\textbf{Positivity:} Indicate how positive the reframed sentence is.

\textbf{1:} As negative as the original sentence.

\textbf{3:} Neutral Sentiment, i.e. neither negative nor positive.

\textbf{5:} Very positive compared to the original sentence. 

Choose \textbf{2} or \textbf{4} when you are hesitant.

\textbf{Fluency:} Indicate the fluency of the reframed sentence.

\textbf{1:} The reframed sentence does not make sense and it is unreadable.

\textbf{3:} The reframed sentence contains some minor grammatical errors, but does not affect reading.

\textbf{5:} The reframed sentence is human-like, without any grammatical errors.

Choose \textbf{2} or \textbf{4} when you are hesitant.




\section{Additional Results}
\label{fir:appendix}

\subsection{Reframe Strategy Classification}
\label{sec:Reframe strategy classification}
We provide the detailed scores of our models on all classification evaluation metrics (i.e., accuracy, precision, recall, and F1 score) for others to compare and refer to, which can be found in Table \ref{detailclass}.

\subsection{Unconstrained Positive Reframing}
\label{sec:Unconstrained Positive Reframing}
For this task, we provide additional ablation results of unconstrained positive reframing in Tables \ref{unconsT5} and \ref{unconsBART}. It can be seen that when the positive sentiment reward is not used, the model's score on metrics such as $\Delta$TB and RTQE decrease. And when the content preservation reward is not used, the model's performance on metrics such as ROUGE and BLEU may decline. In addition, it can be found that the improvement brought by multi-dimensional re-ranking is tremendous, significantly improving the model performance on multiple metrics, indicating that it can better select sentences that meet the requirements of positive reframing from the candidate text set. Based on the above experimental results and analysis, the effectiveness and rationality of each component of MSOF can be fully demonstrated.

\subsection{Controlled Positive Reframing}
\label{sec:Controlled positive reframing}
Similar to Appendix \ref{sec:Unconstrained Positive Reframing}, we provide more detailed ablation experimental results of controlled positive reframing in Tables \ref{controT5} and \ref{controBART}. In addition to the conclusions already drawn in the unconstrained setting, it can be observed that beam search generates sentences with higher content preservation and achieves great results on ROUGE and BLEU. On the other hand, random sampling strategies, namely Top-k, Top-p, and Typical may yield lower scores on ROUGE and BLEU, but achieve better results on $\Delta$TB, RTQE, and PPL, indicating that their generated text may not overlap much with human reference, but still aligns the task requirements and people's daily usage habits better. This is also an important reason why we propose the RTQE metric, which can directly evaluate the degree of reframing of the model-generated text on the original text, thereby avoiding problems caused by unique human reference.

Additionally, we present the ablation results of multi-dimensional re-ranking under controlled setting in Table \ref{detailCon}. It can be observed that when the strategy consistency evaluation is not used, the scores of MSOF$_{{\rm Top-k}}$ on RTQE and PPL  will decrease significantly, but it has better performance on ROUGE and BLEU. When the text similarity evaluation is not used, the performance of MSOF$_{{\rm Top-k}}$ would significantly lower on content preservation-related metrics, but achieves best or sub-optimal results on $\Delta$TB and RTQE. And when the fluency evaluation is not used, the model scores significantly lower on PPL, but still achieves sub-optimal results on RTQE and content preservation-related metrics. This paper suggests that the reason for the above phenomenon may be that the strategy consistency evaluation considers excessive content preservation as indicating incomplete reframing, and thus interacts with the text similarity evaluation. In addition, as can be seen from the results in the table, a decrease in text fluency (high PPL) is often accompanied by a decrease on $\Delta$TB and RTQE. Therefore, there may be some positive correlation among them. Finally, although the overall framework does not achieve optimal results on all metrics, considering the performance of each variant model on each metric, choosing this way is the best trade-off.
\subsection{Positive Reframing under ICL}
\label{ICL}
In this section,  we test the experimental results of GPT-3.5 on the unconstrained positive reframing task and compare it with MSOF\footnote{We use GPT-3.5-turbo-0613 for the experiment. And due to the difficulty of understanding the true meaning of the reframing strategy in LLM under In-Context Learning (ICL), we only conduct experiments under unconstrained setting.}. Under zero-shot, we use \textit{Rephrase the above sentence to be more positive} \citep{ziems2024can} as the instruction. And under few-shot, following \citep{sharma2023cognitive, ziems-etal-2022-inducing}, we retrieve 5 representative exemplars with the closest semantic similarity from the training set as the context for each test case.

As shown in Table \ref{ICLExe}, MSOF still outperforms LLM regarding ROUGE, BLEU, BScore, PPL, and $\Delta$TB, but LLM achieves higher RTQE scores. Based on the examples in Table \ref{unconexample}, it can be seen that although the output of LLM reflects the concept of reframing, it tends to generate longer text and is more prone to hallucinations. This demonstrates the applicability of MSOF for positive reframing and the continued significance of studying small models in this task.

\subsection{Case Study}
\label{sec:appendix}
We provide the generated examples of unconstrained and controlled experiments in Tables \ref{unconexample} and \ref{example}.  A comparative analysis reveals that our models generate more diverse and comprehensive outputs while effectively preserving the underlying meaning of the original text. Specifically, the outputs of the BART-based models are mostly similar, except for the sentences generated by Typical sampling. On the other hand, the T5-based models outperform the BART-based models and baselines by providing the benefits of weekends consistent with human reference. Additionally, although the text in the dataset may contain colloquialisms and even grammatical errors, our models can generate more formal sentences that avoid these issues. Therefore, we speculate that further cleaning and filtering of the data in the dataset can further improve the model's performance. By comparing the results generated by the model in the unconstrained and controlled settings, it can be inferred that without reframe strategy, the reframing performance of the models will decrease, which proves that the reframing strategy plays an auxiliary role in helping the model generate results that better meet task requirements.  

Finally, to further explore whether different reframe strategy will affect the generation results of the model, Table \ref{SameStrategy} shows the generation result of using different strategy to reframe the same negative text. It is obvious from the results that the model can generate reframing text with corresponding characteristics under the guidance of different reframe strategy, especially "Self-affirmation", "Thankfulness" and "Growth Mindset". This proves that the model can learn some information from the reframe strategy and it also shows that the research on controlled positive reframing is valuable.

\begin{table*}[h]
\centering
\begin{tabular}{ccccccccc}
\toprule
\multirow{2}{*}{\textbf{Label}} & \multicolumn{4}{c}{\textbf{Strategy-BERT}} &  \multicolumn{4}{c}{\textbf{Strategy-RoBERTa}}\\
 \cmidrule(lr){2-5} \cmidrule(lr){6-9}& P(\%) & R(\%) & F1(\%) & Acc(\%) & P(\%) & R(\%) & F1(\%) & Acc(\%) \\
\midrule
Thankfulness     & 77.55 & 69.72 & \textbf{73.43} & \textbf{93.41} & 76.84 & 66.97 & 71.57 & 93.05\\
Neutralizing     & 52.75 & 72.84 & \textbf{61.20} & 66.59 & 58.70 & 62.58 & 60.58 & \textbf{70.54}\\
Optimism     & 61.04 & 85.00 & 71.06 & 66.83 & 63.57 & 84.50 & \textbf{72.69} & \textbf{69.58}\\
Impermanence     & 56.10 & 58.60 & \textbf{57.32} & 83.59 & 49.76 & 65.61 & 56.59 & 81.08 \\
Growth Mindset & 58.70  & 77.82 & 66.92  & 79.64 & 65.04  & 72.40 & \textbf{68.52} & \textbf{82.40}\\
Self-affirmation     & 50.72  & 46.05 & \textbf{48.28}  & \textbf{91.02} & 47.22 & 44.74 & 45.94 & 90.42\\
\bottomrule
\end{tabular}
\caption{The detailed experimental results of reframe strategy classification. We provide detailed experimental results of our models on all classification metrics here for analysis and comparison. And the best results in each label are in \textbf{bold}.}
\label{detailclass}
\end{table*}













\begin{table*}[htbp]
\centering

\begin{tabular}{l|cccccccc}
\toprule
\textbf{Model}  & \textbf{R-1} & \textbf{R-2} & \textbf{R-L} & \textbf{BLEU} & \textbf{BScore} & \textbf{$\Delta$TB} & \textbf{RTQE} & \textbf{PPL}    \\
\midrule

MSOF$_{{\rm Greedy}}$ & 32.9 & 13.0 & 26.0 & 8.8 & 89.1 & 0.37 & 86.2 & 36.8\\

~~ w.o Cls & 32.3 & 12.9 & 25.8 & 8.8 & 89.1 & 0.37 & 86.1 & 39.6\\

~~ w.o Cont & 32.6 & 12.7 & 25.7 & 8.4 & 89.0 & 0.38 & 87.6 & 38.5\\ \midrule

MSOF$_{{\rm Beam}}$ & 34.1 & 14.0 & 27.1 & 9.7 & 89.2 & 0.37 & 89.0 & 35.4\\ 

~~ w.o Cls & 33.6 & 13.7 & 26.8 & 9.5 & 89.2 & 0.35 & 88.6 & 36.3\\

~~ w.o Cont & 33.6 & 13.6 & 26.7 & 9.3 & 89.1 & 0.36 & 90.2 & 40.0\\

~~ w.o Re-ranking & 33.1 & 13.2 & 26.3 & 9.1 & 89.1 & 0.36 & 84.3 & 39.5\\ \midrule

MSOF$_{{\rm Top-k}}$ &  
34.8 & 14.7 & 27.7 & 10.1 & 89.5 & 0.44 & 93.5 & 22.3\\

~~ w.o Cls & 34.6 & 14.9 & 27.8 & 10.2 & 89.5 & 0.42 & 93.5 & 22.6\\

~~ w.o Cont & 34.0 & 14.5 & 27.4 & 9.6 & 89.4 & 0.39 & 94.1 & 23.6\\

~~ w.o Re-rankig & 31.9 & 11.7 & 25.1 & 7.7 & 89.1 & 0.42 & 92.7 & 27.0\\ \midrule

MSOF$_{{\rm Top-p}}$ & 34.4 & 14.6 & 27.6 & 10.1 & 89.4 & 0.43 & 93.5 & 22.2\\

~~ w.o Cls & 34.6 & 14.6 & 27.6 & 9.9 & 89.5 & 0.42 & 94.1 & 23.4\\

~~ w.o Cont & 34.2 & 14.6 & 27.6 & 9.7 & 89.4 & 0.37 & 93.6 & 21.5\\

~~ w.o Re-ranking & 31.9 & 12.2 & 25.3  & 8.2  & 89.1 & 0.41 & 90.7 & 28.0\\ \midrule

MSOF$_{{\rm Typical}}$ & 32.9 & 13.5 & 26.2 & 9.1 & 89.3 & 0.39 & 94.5 & 22.6\\

~~ w.o Cls & 33.4 & 13.6 & 26.7 & 8.9 & 89.3 & 0.42 & 95.7 & 22.8\\

~~ w.o Cont & 32.2 & 12.9 & 25.8 & 8.3 & 89.2 & 0.38 & 95.3 & 23.0\\

~~ w.o Re-ranking & 31.7 & 12.0 & 25.3  & 8.0  & 89.1 & 0.40 & 92.2 & 31.7\\

\bottomrule
\end{tabular}
\caption{The detailed experimental results of \textbf{unconstrained positive reframing} (\textbf{T5}). }
\label{unconsT5}
\end{table*}

\begin{table*}[htbp]
\centering

\begin{tabular}{l|cccccccc}
\toprule
\textbf{Model}  & \textbf{R-1} & \textbf{R-2} & \textbf{R-L} & \textbf{BLEU} & \textbf{BScore} & \textbf{$\Delta$TB} & \textbf{RTQE} & \textbf{PPL}    \\
\midrule

MSOF$_{{\rm Greedy}}$ & 32.3 & 13.2 & 26.9 & 10.4 & 89.4 & 0.24 & 80.1 & 47.0 \\

~~ w.o Cls & 32.9 & 13.3 & 27.2 & 10.1 & 89.3 & 0.20 & 75.9 & 53.7\\

~~ w.o Cont & 32.4  & 13.0 & 26.8 & 10.3 & 89.2 & 0.26 & 79.7 & 63.0\\ \midrule

MSOF$_{{\rm Beam}}$ & 34.2 & 14.2 & 28.1 & 10.9 & 89.5 & 0.24 & 87.3 & 33.6\\

~~ w.o Cls & 34.1 & 14.2 & 27.9 & 10.6 & 89.5 & 0.22 & 85.9 & 35.0\\

~~ w.o Cont & 33.6 & 13.8 & 27.7 & 10.6 & 89.4 & 0.30 & 86.1 & 36.0\\

~~ w.o Re-ranking & 33.3 & 13.5 & 27.4 & 10.3 &  89.5 & 0.29 & 88.0 & 44.8\\ \midrule

MSOF$_{{\rm Top-k}}$ &  34.8 & 14.9 & 29.3 & 12.0 & 89.9 & 0.31 & 87.3 & 25.8\\

~~ w.o Cls & 34.9 & 15.1 & 29.1 & 12.2 & 89.8 & 0.31 & 85.6 & 30.2\\

~~ w.o Cont & 34.7 & 15.0 & 29.0 & 12.2 & 89.8 & 0.27 &  84.1 & 30.5\\

~~ w.o Re-ranking & 31.6 & 11.7 & 26.0 & 9.4 & 89.4 & 0.28 & 84.8 & 38.9\\ \midrule

MSOF$_{{\rm Top-p}}$ & 34.8 & 14.9 & 29.2 & 12.0 & 89.8 & 0.30 & 87.2 & 27.3\\

~~ w.o Cls & 34.4 & 14.4 & 28.5 & 11.5 & 89.7 & 0.27 & 84.2 & 31.6\\

~~ w.o Cont & 34.8 & 14.8 & 29.0 & 11.8 & 89.8 & 0.28 &  86.1 & 31.5\\

~~ w.o Re-ranking & 31.4 & 11.9 & 25.9 & 9.3 & 89.4 & 0.29 & 85.6 & 37.3\\ \midrule

MSOF$_{{\rm Typical}}$ & 32.5 & 12.8 & 26.9 & 10.4 & 89.5 & 0.30 & 88.5 & 29.6\\

~~ w.o Cls & 32.6 & 13.2 & 26.9 & 10.8 & 89.5 & 0.28 & 87.1 & 32.6\\

~~ w.o Cont & 33.0 & 13.2 & 27.3 & 10.7 & 89.5 & 0.34 &  92.8 & 32.6\\

~~ w.o Re-ranking & 31.5 & 11.9 & 25.8 & 9.2 & 89.2 & 0.25 & 82.4 & 41.3\\

\bottomrule
\end{tabular}
\caption{The detailed experimental results of \textbf{unconstrained positive reframing} (\textbf{BART}). }
\label{unconsBART}
\end{table*}

\begin{table*}[htbp]
\centering
\begin{tabular}{l|ccccccccc}
\toprule
 \textbf{Model} &  \textbf{R-1} & \textbf{R-2} & \textbf{R-L} & \textbf{BLEU} & \textbf{BScore} & \textbf{$\Delta$TB} & \textbf{RTQE} & \textbf{PPL}    \\
\midrule

MSOF$_{{\rm Greedy}}$ & 33.6 & 13.6 & 26.7 & 8.8 & 89.2 & 0.37 & 94.6 & 34.6\\ 

~~ w.o Cls & 33.5 & 13.4 & 26.6 & 8.9 & 89.1 & 0.35 & 91.2 & 38.4\\

~~ w.o Cont & 33.3 & 13.2 & 26.3 & 8.6 & 89.2 & 0.32 & 88.8 & 41.2\\ \midrule

MSOF$_{{\rm Beam}}$ & 34.6 & 14.4 & 27.5 & 9.5 & 89.3 & 0.36 & 96.2 & 34.5\\  

~~ w.o Cls & 33.7 & 13.7 & 26.5 & 8.9 & 89.2 & 0.30 & 94.3 & 39.7\\

~~ w.o Cont  & 33.7 & 13.6  & 26.7 & 9.1  & 89.1 & 0.34 & 89.1 & 40.3\\

~~ w.o Re-ranking &  33.9 & 13.7 & 26.9 & 9.1 & 89.2 & 0.36 & 93.0 & 37.3\\ \midrule

MSOF$_{{\rm Top-k}}$ & 34.8 & 15.0 & 28.0 & 9.9 & 89.5 & 0.43 & 97.7 & 23.1\\

~~ w.o Cls & 34.5 & 14.5 & 27.5 & 9.4 & 89.4 & 0.41 & 96.7 & 25.3\\

~~ w.o Cont  & 35.0 & 14.8  & 27.7 & 9.6  & 89.6 & 0.37 & 95.7 & 24.2\\

~~ w.o Re-ranking & 32.1 & 12.0 & 25.2 & 7.6 & 89.1 & 0.43 & 96.1 & 28.3\\ \midrule

MSOF$_{{\rm Top-p}}$ & 34.1 & 14.2 & 27.6 & 9.3 & 89.5 & 0.42 & 96.6 & 23.0\\

~~ w.o Cls & 34.6 & 14.5 & 27.5 & 9.5 & 89.5 & 0.41 & 95.7 & 25.7\\

~~ w.o Cont  & 34.3 & 14.2  & 27.2 & 9.2  & 89.5 & 0.39 & 95.7 & 27.7\\

~~ w.o Re-ranking & 32.1 & 11.8 & 25.4 & 7.3 & 89.2 & 0.43 & 95.3 & 28.0\\ \midrule

MSOF$_{{\rm Typical}}$ & 33.2 & 13.4 & 26.5 & 8.6 & 89.3 & 0.42 & 97.0 & 23.8\\

~~ w.o Cls & 33.2 & 13.2 & 26.3 & 8.5 & 89.3 & 0.41 & 96.9 & 26.2\\

~~ w.o Cont  & 33.2 & 13.7  & 26.2 & 8.9  & 89.4 & 0.37 & 97.4 & 25.3\\

~~ w.o Re-ranking & 32.3 & 12.2 & 25.5 & 7.7 & 89.2 & 0.42 & 95.3 & 28.3\\
\bottomrule
\end{tabular}
\caption{The detailed experimental results of \textbf{contolled positive reframing} (\textbf{T5}).}
\label{controT5}
\end{table*}

\begin{table*}[htbp]
\centering
\begin{tabular}{l|ccccccccc}
\toprule
 \textbf{Model} &  \textbf{R-1} & \textbf{R-2} & \textbf{R-L} & \textbf{BLEU} & \textbf{BScore} & \textbf{$\Delta$TB} & \textbf{RTQE} & \textbf{PPL}    \\
\midrule

MSOF$_{{\rm Greedy}}$ & 33.0 & 13.3 & 27.2 & 10.0 & 89.6 & 0.31 & 89.1 & 44.4\\

~~ w.o Cls  & 31.8 & 12.7 & 26.6 & 10.2 & 89.5 & 0.27 & 82.3 & 57.0\\

~~ w.o Cont  & 33.3 & 13.2 & 27.0 & 9.8  & 89.5 & 0.29 & 88.6 & 47.4\\  \midrule

MSOF$_{{\rm Beam}}$ & 34.6 & 14.2 & 28.2 & 10.5 & 89.7 & 0.34 & 94.8 & 31.8 &\\

~~ w.o Cls  & 33.8 & 14.2 & 27.9 & 10.8 & 89.6 & 0.30 & 90.7 & 36.4\\

~~ w.o Cont  & 35.1 & 14.5 & 28.3 & 10.3  & 89.6 & 0.33 & 94.1 & 33.\\

~~ w.o Re-rank & 33.2 & 13.5 & 27.5 & 10.3 & 89.5 & 0.29 & 88.0 & 44.8\\ \midrule

MSOF$_{{\rm Top-k}}$ & 34.8 & 14.7 & 29.0 & 11.4 & 90.1 & 0.36 & 94.0 & 29.4\\

~~ w.o Cls  & 33.6 & 13.7 & 28.2 & 10.8 & 90.0 & 0.35 & 86.9 & 31.3\\

~~ w.o Cont  & 33.1 & 13.7 & 27.5 & 10.9 & 89.7 & 0.38 & 86.2 & 34.6\\

~~ w.o Re-ranking & 31.9 & 11.9 & 26.2 & 9.4 & 89.6 & 0.35 & 92.9 & 38.8\\ \midrule

MSOF$_{{\rm Top-p}}$ & 34.6 & 14.4 & 28.8 & 11.3 & 90.0 & 0.36 & 94.0 & 30.8\\

~~ w.o Cls  & 34.0 & 14.2 & 28.4 & 10.8 & 90.0 & 0.35 & 88.1 & 34.0\\

~~ w.o Cont  & 33.5 & 14.0 & 27.9 & 11.2  & 89.8 & 0.39 & 87.8 & 33.6\\

~~ w.o Re-ranking & 31.4 & 11.9 & 26.2 & 8.9 & 89.6 & 0.33 & 86.9 & 47.1\\ \midrule

MSOF$_{{\rm Typical}}$ & 33.2 & 13.2 & 27.5 & 10.1 & 89.8 & 0.36 & 94.0 & 29.8\\

~~ w.o Cls  & 32.2 & 12.6 & 26.8 & 9.5 & 89.7 & 0.34 & 86.4 & 38.9\\

~~ w.o Cont  & 32.1 & 12.6 & 26.4 & 10.0  & 89.5 & 0.36 & 90.3 & 37.6\\

~~ w.o Re-ranking & 31.0 & 11.6 & 25.7 & 8.7 & 89.5 & 0.34 & 86.5 & 45.6\\

\bottomrule
\end{tabular}
\caption{The detailed experimental results of \textbf{contolled positive reframing} (\textbf{BART}).}
\label{controBART}
\end{table*}

\begin{table*}[htbp]
\centering
\begin{tabular}{llcccccccc}
\toprule
 \multicolumn{2}{c}{\textbf{Model}} &  \textbf{R-1} & \textbf{R-2} & \textbf{R-L} & \textbf{BLEU} & \textbf{BScore} & \textbf{$\Delta$TB} & \textbf{RTQE} & \textbf{PPL}    \\
\midrule

\multirow{4}{*}{\textbf{T5}}

&MSOF$_{{\rm Top-k}}$ & 34.8 & 15.0 & 28.0 & 9.9 & \textbf{89.5} & 0.43 & \textbf{97.7} & 23.1\\

& ~~w.o Strategy & \textbf{35.6} & \textbf{15.8} & \textbf{28.8} & \textbf{10.7} & 89.5 & 0.41 & 95.0 & 30.0\\

& ~~w.o Similar & 32.2 & 12.1 & 25.4 & 7.6 & 89.2 & \textbf{0.44} & 97.5 & \textbf{21.3} \\

& ~~w.o Fluency & 35.0 & 15.3 & 28.1 & 10.1 & 89.5 & 0.41 & 97.1 & 28.6 \\

\hline

\multirow{4}{*}{\textbf{BART}}

& MSOF$_{{\rm Top-k}}$ & \textbf{34.8} & \textbf{14.7} & \textbf{29.0} & 11.4 & \textbf{90.1} & 0.36 & \textbf{94.0} & \textbf{29.4}\\

& ~~w.o Strategy & 34.0 & 14.6 & 28.4 & \textbf{11.8} & 89.7 & 0.37 & 84.3 & 34.0 \\

& ~~w.o Similar & 29.6 & 10.6 & 24.4 & 8.3 & 89.3 & \textbf{0.41} & 85.8 & 32.3 \\

& ~~w.o Fluency & 33.9 & 14.3 & 28.2 & 11.6 & 89.7 & 0.35 & 86.2 & 46.9 \\

\bottomrule
\end{tabular}
\caption{The ablation experimental results of \textbf{multi-dimensional re-ranking}. w.o Strategy means without strategy consistency evaluation, w.o Similarity represents without textual similarity evaluation and w.o Fluency represents not using fluency evaluation.}
\label{detailCon}
\end{table*}

\begin{table*}[htbp]
\centering
\begin{tabular}{l|ccccccccc}
\toprule
 \textbf{Setting} &  \textbf{R-1} & \textbf{R-2} & \textbf{R-L} & \textbf{BLEU} & \textbf{BScore} & \textbf{$\Delta$TB} & \textbf{RTQE} & \textbf{PPL}    \\
\midrule

Zero-shot & 27.0 & 7.2 & 21.2 & 4.2 & 88.9 & 0.43 & 99.8 & 42.0\\

Five-shot & 28.5 & 8.5 & 22.5 & 5.3 & 89.3 & 0.28 & 97.1 & 27.5 \\ 

\bottomrule
\end{tabular}
\caption{The performance of GPT3.5 on positive reframing.}
\label{ICLExe}
\end{table*}

\begin{table*}[htbp]
\centering
\resizebox{\linewidth}{!}{
\begin{tabular}{ll}
\toprule
\textbf{Original text}  & So glad that tomorrow is Friday. This has seriously been the longest week of my life\\
\midrule
\textbf{Reference} & I'm glad the weekend is coming up, \colorbox{pink}{so I can rest}.\\ \midrule

T5 \citep{ziems-etal-2022-inducing} & This week has been a long one, but I'm sure it will be over soon.\\

FDSC \cite{Sheng-Decoupling} & I'm so glad that tomorrow is Friday. This week has been a long one.\\

ST2PG \cite{sheng-etal-2023-learning} & I'm glad that tomorrow is Friday. This has been the longest week of my life. \\

PG2ST \cite{sheng-etal-2023-learning} & I'm glad that tomorrow is Friday. This has been the longest week of my life. \\

MSOF$_{{\rm Beam}}$ & I'm glad that tomorrow is Friday. It's been a long week, \colorbox{pink}{but it's going to be a good one}.\\

MSOF$_{{\rm Top-k}}$ & This week has been a long week, but I'm glad it's Friday. \colorbox{pink}{I'll be able to relax and enjoy the weekend}.\\

MSOF$_{{\rm Top-p}}$  & It's been a long week, \colorbox{pink}{but it's a good chance to get some rest}.\\

MSOF$_{{\rm Typical}}$ & I'm glad that tomorrow is Friday. This week has been challenging, but I'm going to get through it.\\ \midrule

BART \citep{ziems-etal-2022-inducing}  & I'm glad that tomorrow is Friday. This has been the longest week of my life, but I'm sure I'll get through it.\\

FDSC \cite{Sheng-Decoupling} & So glad that tomorrow is Friday. This has been the longest week of my life. I'm tired, but I'm sure I can get through it.\\

ST2PG \cite{sheng-etal-2023-learning} & I'm glad that tomorrow is Friday. This has been the longest week of my life, but I'm sure it will be over soon. \\

PG2ST \cite{sheng-etal-2023-learning} & I'm glad that tomorrow is Friday. This has been the longest week of my life, but I'm sure it will be over soon. \\

MSOF$_{{\rm Beam}}$ & I'm glad that tomorrow is Friday. This week has been very challenging.\\

MSOF$_{{\rm Top-k}}$  & I'm glad that tomorrow is Friday. This week has been very challenging.\\

MSOF$_{{\rm Top-p}}$  & I'm glad that tomorrow is Friday. This week has been so long.\\

MSOF$_{{\rm Typical}}$  & I'm glad that tomorrow is Friday. This week has been challenging, but I'm going to get through it.\\
\midrule
GPT-3.5 (zero-shot) & I'm excited that tomorrow is finally Friday! \colorbox{pink}{This week has been full of new experiences and opportunities}. \\

GPT-3.5 (five-shot) & I can't wait for Friday to finally arrive! This week has been a challenge, but I made it through. tired\\

\bottomrule
\end{tabular}}
\caption{The reframing examples of \textbf{unconstrained positive reframing}. In order to better compare with the constrained settings.  The \colorbox{pink}{pink text} shows the positive perspective.}
\label{unconexample}
\end{table*}

\begin{table*}[htbp]
\centering
\resizebox{\linewidth}{!}{
\begin{tabular}{ll}
\toprule
\textbf{Original text} & So glad that tomorrow is Friday. This has seriously been the longest week of my life! tired \\
\midrule

\textbf{Reference} & I’m glad the weekend is coming up, \colorbox{pink}{so I can rest}. \\\midrule

T5 \citep{ziems-etal-2022-inducing} & So glad that tomorrow is Friday. This has seriously been the longest week of my life. I'm tired, but I know I'll get through it.\\

MSOF$_{{\rm Beam}}$ & I'm so glad that tomorrow is Friday. This has been the longest week of my life, \colorbox{pink}{but I know that tomorrow will be a better day}.\\

MSOF$_{{\rm Top-k}}$  & I'm glad that tomorrow is Friday. This has been the longest week of my life, \colorbox{pink}{and I've had a lot of fun}.\\

MSOF$_{{\rm Top-p}}$  & I'm glad that tomorrow is Friday. This has been the longest week of my life, \colorbox{pink}{but I know it's going to be a great day.}\\

MSOF$_{{\rm Typical}}$  & Tomorrow is Friday. This has been the longest week of my life, but I know I will make it to the end of the week. \colorbox{pink}{It will be great.}\\\midrule

BART \citep{ziems-etal-2022-inducing} & I’m glad that tomorrow is Friday. This week has been long, but I’m looking forward to the weekend.\\

MSOF$_{{\rm Beam}}$  & I'm so glad that tomorrow is Friday. This has been the longest week of my life! I'm tired \colorbox{pink}{but I'm sure it will be good}.\\ 

MSOF$_{{\rm Top-k}}$  & I'm really looking forward to Friday, \colorbox{pink}{so I can relax a bit}.\\ 

MSOF$_{{\rm Top-p}}$  & I'm glad that tomorrow is Friday. \colorbox{pink}{I'm going to feel so much better}. \\ 

MSOF$_{{\rm Typical}}$  & Even though Friday is the longest week in my life, \colorbox{pink}{I'm happy to have the chance to rest for a few days}.\\ 

\bottomrule
\end{tabular}}
\caption{The model comparison for reframing the same text and the reframing strategy is optimism.  And we selected the same example as Table \ref{unconexample} to better compare the output of models under different settings.}
\label{example}
\end{table*}

\begin{table*}[htbp]
\centering
\resizebox{\linewidth}{!}{
\begin{tabular}{lll}
\toprule
\multicolumn{2}{l}{\textbf{Original text}} & I hate that I stress my self out so much that I can't fall asleep! \\
\midrule

\multirow{6}{*}{MSOF$_{{\rm Top-k}}$(T5)}

& Growth Mindset & \makecell[l]{I need to take better care of myself so that I can fall asleep in no time! I'm going to try to reduce my stress and \\ improve my sleep.}\\

& Impermanence & I don't like that I stress myself out so much that I can't fall asleep, but I'm sure I'll get better soon.\\

& Neutralizing & I am stressed out so much that I can't fall asleep, but I'm going to take a nap and sleep better so I can sleep better.\\

& Optimism & I don't like to stress myself out so much that I can't fall asleep, but I'm sure I'll fall asleep soon.\\

& Self-affirmation & I don't like that I stress my self out so much that I can't fall asleep, but I'm a strong person, and I know I can do it.\\

& Thankfulness & I'm glad I have a bed to sleep in after a long day of stressing myself out, I can't sleep.\\

\midrule

\multirow{6}{*}{MSOF$_{{\rm Top-k}}$(BART)}

& Growth Mindset & I'm going to stop stressing out about things so that I can fall asleep.\\

& Impermanence & I'm going to take some time to myself to clear my head.\\

& Neutralizing & Stress is part of life, and I can't fall asleep, but I'm sure I'll feel better soon.\\

& Optimism & I'm going to have to stay up all night tonight so that I can get some peace of mind.\\

& Self-affirmation & I am not able to sleep because of my stress. But I am a strong person, and I know I can get through this.\\

& Thankfulness & I'm thankful that I have a bed to sleep in when I'm stressed.\\

\bottomrule
\end{tabular}}
\caption{A model comparison for reframing the same text using different reframe strategy.}
\label{SameStrategy}
\end{table*}

\end{document}